\documentclass[10pt,twocolumn,letterpaper]{article}

\usepackage{cvpr}              

\usepackage{graphicx}
\usepackage{amsmath}
\usepackage{amssymb}
\usepackage{booktabs}

\usepackage[pagebackref,breaklinks,colorlinks]{hyperref}

\usepackage[capitalize]{cleveref}
\crefname{section}{Sec.}{Secs.}
\Crefname{section}{Section}{Sections}
\Crefname{table}{Table}{Tables}
\crefname{table}{Tab.}{Tabs.}

\def\cvprPaperID{8198} 
\def\confName{CVPR}
\def\confYear{2022}

\begin{document}

\title{FashionSearchNet-v2: Learning Attribute Representations with Localization for Image Retrieval with Attribute Manipulation}

\author{Kenan E. Ak\textsuperscript{1} \hspace{0.1cm} Joo Hwee Lim\textsuperscript{1}  \hspace{0.1cm} Ying Sun\textsuperscript{1} \hspace{0.1cm} Jo Yew Tham\textsuperscript{2}  \hspace{0.1cm} Ashraf A. Kassim\textsuperscript{3}  \\
\textsuperscript{1}Institute for Infocomm Research, A*STAR, Singapore   \\ 
\textsuperscript{2}ESP xMedia Pte. Ltd., Singapore   \\ 
\textsuperscript{3}Singapore University of Technology and Design, Singapore  \\ 
{\tt\scriptsize	kenanea@i2r.a-star.edu.sg, joohwee@i2r.a-star.edu.sg, suny@i2r.a-star.edu.sg, thamjy@espxmedia.com, ashraf@nus.edu.sg}
}
\maketitle


\begin{abstract}
The focus of this paper is on the problem of image retrieval with attribute manipulation. Our proposed work is able to manipulate the desired attributes of the query image while maintaining its other attributes. For example, the collar attribute of the query image can be changed from round to v-neck to retrieve similar images from a large dataset. A key challenge in e-commerce is that images have multiple attributes where users would like to manipulate and it is important to estimate discriminative feature representations for each of these attributes. The proposed FashionSearchNet-v2 architecture is able to learn attribute specific representations by leveraging on its weakly-supervised localization module, which ignores the unrelated features of attributes in the feature space, thus improving the similarity learning. The network is jointly trained with the combination of attribute classification and triplet ranking loss to estimate local representations. These local representations are then merged into a single global representation based on the instructed attribute manipulation where desired images can be retrieved with a distance metric. The proposed method also provides explainability for its retrieval process to help provide additional information on the attention of the network. Experiments performed on several datasets that are rich in terms of the number of attributes show that FashionSearchNet-v2 outperforms the other state-of-the-art attribute manipulation techniques. Different than our earlier work (FashionSearchNet), we propose several improvements in the learning procedure and show that the proposed FashionSearchNet-v2 can be generalized to different domains other than fashion.
\end{abstract}

\section{Introduction}
Over the recent years, there has been remarkable progress in fashion related research including attribute/object recognition \cite{Dong,Liu,simo2016fashion,li2019two}, attribute discovery \cite{han2017automatic,hsiao2017learning,vittayakorn2016automatic}, recommendation \cite{al2017fashion,han2017learning,jing2019low,hsiao2018creating,kang2019complete,sattar2019fashion,singhal2020towards,zhan2021a3}, human/fashion parsing \cite{liang2015human,Simo-serra,yamaguchi2013paper}, retrieval \cite{corbiere2017leveraging,hadi2015buy,vo2019composing,ak2018shirt,lang2020plagiarism}. Furthermore, with the success of generative networks \cite{goodfellow2014generative,salimans2017pixelcnn++,karras2020analyzing,zhu2017unpaired,choi2017stargan,arjovsky2017wasserstein,ak2020incorporating,zou2020edge}, an extensive research is being conducted that involves synthesis of fashion images synthesis \cite{han2019finet,han2018viton,ak2019attribute,kenan_ICIP,ak2020semantically,zhu2017your,cui2021dressing}. Among these wide range of problems, the focus of this paper is on image retrieval. While the key issue addressed in the mainstream image retrieval research is on cross-domain image retrieval, a major challenge is managing situations when there is a large number of attributes that can be used for attribute manipulation of the query image, which requires flexible \& discriminative representations in the retrieval systems.

Conducting an image search by changing a certain attribute of the query image is tricky, since it may be hard to find the right balance between ``maintaining the current attributes" and ``adding/replacing an attribute". In \cite{zhaobo_atman}, this problem is defined as image retrieval with attribute manipulation and Attribute Manipulation Network (AMNet) is proposed to address the problem. 
AMNet, trained with triplet ranking loss, fuses features of the query image with the desired attribute representation to find the desired image. Another approach involves allowing users to decide which image is more preferred through ``relevance feedback" but it can be computationally intensive \cite{Kovashka2012, Li2016, Tong2000}. In any case, these methods do not explore attribute localization to estimate feature representations, which could help to leave out some artifacts of the unwanted attributes.

\begin{figure}[t]
\centering
\includegraphics[width=8.9cm]{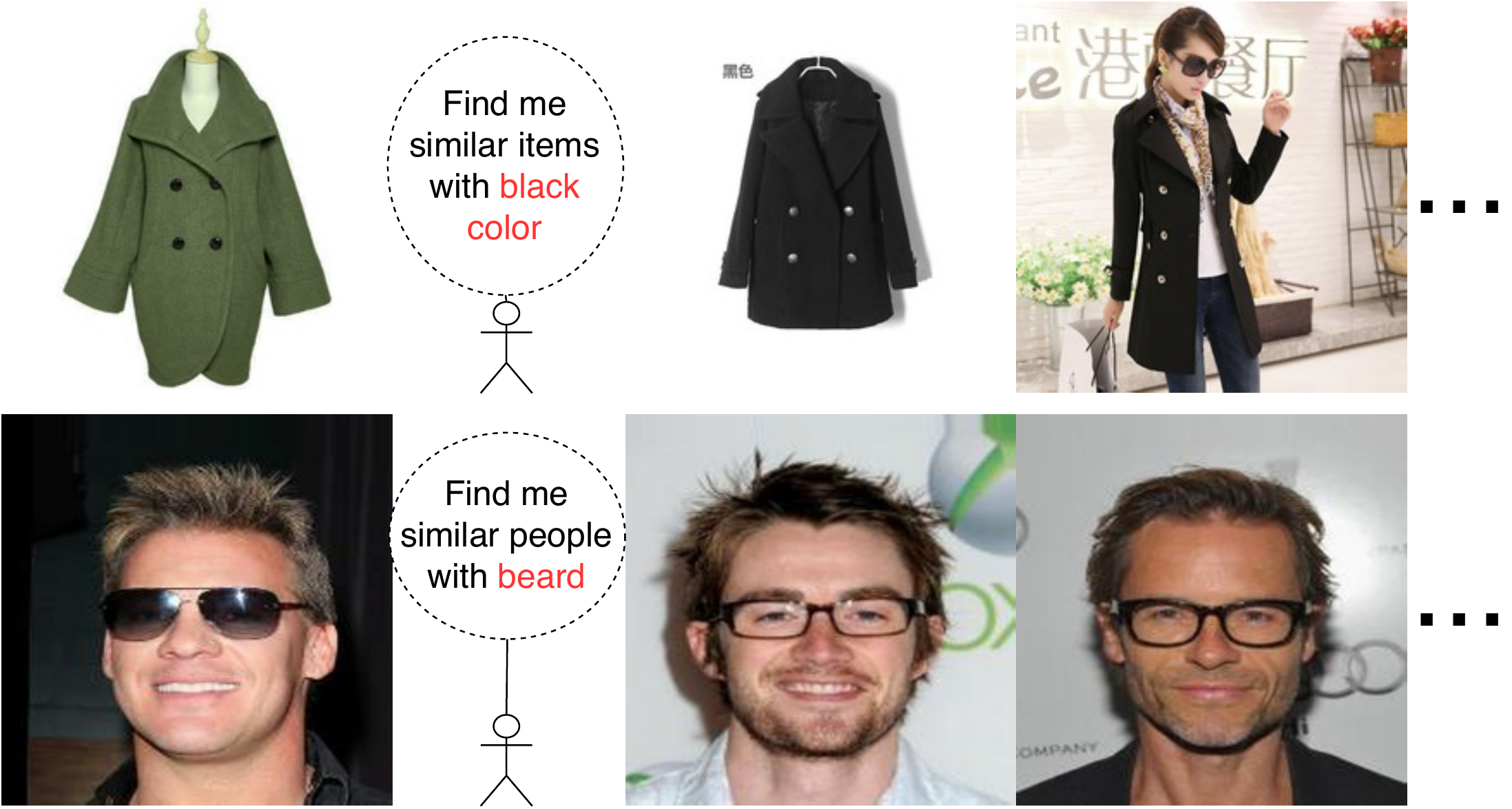}
\caption{For each given query image, the user is able to manipulate specific attributes. Each attribute manipulation (shown with the \textcolor{red}{red font}) redefines the representation of the query image based on the desired attribute, which is then used by FashionSearchNet-v2 to retrieve the most similar images by using the ``redefined" query image.
}
\label{fig:first_fig}
\end{figure}

In this paper, we introduce FashionSearchNet-v2, an image retrieval network based on a query image, and given an attribute manipulation condition. Figure \ref{fig:first_fig} illustrates two attribute manipulation scenarios for given query images. For the first scenario, the color attribute is changed to ``black" while the other attributes in the query image are maintained. In the second example, we show that the proposed idea can be applied to find similar people with a beard attribute. 


In image retrieval with attribute manipulation, there may be many attributes that the user may want to change where each attribute is located in a different region of the image. Focusing on regions of interest is more feasible for images with multiple attributes since some attributes may only be present at certain locations. In line with this objective, the proposed FashionSearchNet-v2 initially aims to learn attribute similarities by leveraging a weakly supervised localization method: class activation mapping (CAM) \cite{zhou2016learning}, i.e., no information is given about the location of the attributes. The localization module is based on the global average pooling (GAP) layer and used to generate attribute activation maps (AAMs), i.e., spatial attention. Consequently, each attribute can be represented by its most relevant region. Next, the region of interest (ROI) pooling is performed to feed these attribute-relevant features into a set of fully connected layers, which serve as local attribute representations. The classification and triplet losses are used to train FashionSearchNet-v2 and learn attribute representations/similarities. 

Traditionally, employing the triplet loss requires anchor, positive and negative images. When the information about the anchor and positive images is not available, one can choose to select them based on their attribute similarities. However, this may lead to having very few examples in the training set. Therefore, we propose to use ``triplets of regions", which can be computed independently for each individual attribute representation. By disconnecting attribute representations into different layers, the triplet loss for computing similarity between query and candidate images can now be computed on triplets of regions where each region is estimated from the AAMs. Consequently, this process increases the number of possible triplets immensely as the learning is now independent of the attribute quantity. 

After the training of attribute representations is completed, the undesired attribute representation of the query image can be directly replaced for a given attribute manipulation by utilizing the proposed attribute memory block. Next, the network combines attribute representations of the query image into a single global representation and the image retrieval can be performed. In order to allow this, we use the global ranking loss function to teach FashionSearchNet-v2 which representation should have more importance depending on the attribute manipulation.

Compared to the earlier version: FashionSearchNet \cite{ak2018learning}, the proposed FashionSearchNet-v2 includes an updatable attribute memory block to estimate more precise attribute representation prototypes. Additionally, we include a feature fusion method that combines features with and without the localization method. This feature fusion method is especially important when AAMs are inaccurate or there is a dependency among several attributes. We further test our proposed idea on the CelebA dataset \cite{liu2015faceattributes} and show our method can be generalized for different domains for image retrieval with attribute manipulation. We also present more details on the algorithm while providing numerous experiments that include different ablation studies.

The main contributions of the paper can be summarized as follows:
\begin{itemize}
\item We introduce a novel image retrieval with attribute manipulation network: FashionSearchNet-v2 that is able to conduct attribute representation learning by localizing towards attributes and acquire better attribute representations for attribute manipulations.

\item In addition to using attribute activation maps to discover attribute relevant regions, we utilize feature fusion and an updatable memory block to store prototype attribute representations.

\item Our experiments show that each module significantly contributes to the performance of the proposed network while being able to handle different types of domains including fashion and face images \& attributes.
\end{itemize}

%
%
%

\section{Related Work}
\subsection{Attribute Recognition}
Attributes are tools that can be used to describe and categorize objects \cite{sudowe2015person,sarafianos2018deep,siddiquie2011image}. Preliminary works \cite{Bossard2013,Chen,kiapour2014hipster} relied on combining hand-crafted features such as SIFT \cite{Lowe2004} and HoG \cite{Dalal2005} with support vector machines \cite{cortes1995support}. Along with the introduction of deep neural networks, more powerful methods have been proposed for attribute recognition \cite{wang2019pedestrian,tang2019improving,han2019attribute}. 

In terms of fashion products, attributes also provide a useful tool to assess and process clothing items. Mix and match \cite{Yamaguchi} combined a deep network with conditional random fields to explore the compatibility of clothing items and attributes. Chen et al. \cite{Chen2015} focused on solving the problem of describing people based on fine-grained clothing attributes. Several works utilized weakly labeled image-text pairs to discover attributes \cite{vittayakorn2016automatic, yashima2016learning}. Abdulnabi et al. \cite{abdulnabi2015multi} proposed a multi-task based approach to learn an algorithm to predict multi-attributes. Li et al. \cite{li2019two} proposed a two-stream network for fashion recognition. More recently, a technique for hard-aware attribute classification \cite{ye2019hard} is proposed to address the issue of imbalanced data distribution. To enhance the extraction of shape and texture features, a two-stream network \cite{Zhang_2020_CVPR} is proposed. Attribute recognition plays an important part in this work as it is utilized in both localization and retrieval abilities of FashionSearchNet-v2.
\subsection{Attribute Localization}
Being able to localize towards objects has been proven to be important in fine-grained recognition \cite{huang2016part,jaderberg2015spatial,huynh2020fine} and image retrieval \cite{bell2015learning,gordo2016deep}. More specifically, in fashion images, DARN \cite{Huang2015a} utilized from a module to detect clothing items, which subsequently improved the performance of the image retrieval network. Similar to DARN \cite{Huang2015a}, Song et al. \cite{song2017learning} proposed a unified method to localize and classify apparels. More interestingly, in FashionNet \cite{Liu} joint prediction of clothing attributes and landmarks shown to be highly effective.

However, most aforementioned methods require annotation of bounding boxes or key points to correctly detect the object of interest. In \cite{singh2016end}, an end-to-end method is proposed to simultaneously localize and rank relative attributes in a weakly supervised manner with the help of Spatial Transformer Networks (STN) \cite{jaderberg2015spatial}. Nevertheless, the fact that a new model must be trained for each attribute for the method proposed in \cite{singh2016end} makes it hard to implement for images with multiple attributes. In \cite{tang2019improving}, the attribute localization module based on a spatial transformer is shown to improve pedestrian attribute recognition. Another method that allows estimating weakly-supervised localization information is class activation mapping (CAM) \cite{zhou2016learning}, which has been shown to be highly efficient in localizing the most representative regions of attributes. Class activation mapping can be used for many problems such as localizing towards attributes of pedestrians \cite{liu2018localization}, semantic segmentation \cite{wei2017object} and person re-identification \cite{yang2019towards}. As it is not quite possible to annotate bounding boxes for every attribute, we were inspired to innovate by incorporating a weakly annotated attention mechanism to conduct multi-attribute based similarity learning.
\subsection{Image Retrieval}
The success of deep learning based approaches provided better performance for the content-based image retrieval \\(CBIR) \cite{babenko2014neural, wan2014deep, krizhevsky2011using, Huang2015a} compared to the traditional image feature extraction methods. The most popular CBIR system focuses on searching for same/similar items from query images \cite{hadi2015buy,Liu,shankar2017deep,simo2016fashion} or videos \cite{cheng2017video2shop,garcia2017dress}, while another set of works investigate the problem of recommendation \cite{al2017fashion, Liu2012, tangseng2017recommending, veit2015learning, hou2019explainable}, which is also closely related to image retrieval. More recently, retrieval of fashion images is investigated via various methods \cite{Kuang_2019_ICCV,Lin_2020_CVPR,lang2020plagiarism}.

Different than retrieval by the query, there are several works that focus on the problem of adjusting certain aspects of the query image and performing the image search. The user feedback can be incorporated with attribute \cite{han2017automatic,zhaobo_atman,ak2018learning} or text \cite{vo2019composing,Chen_CVPR20,Chen_ECCV20}. Another work \cite{shin2019semi} explores fashion instance-level image retrieval. Han et al. \cite{han2017automatic} focused on spatially aware concept discovery from image/text pairs and conducted attribute-feedback product retrieval with the word2vec \cite{mikolov2013efficient} method. On the other hand, Zhao et al. \cite{zhaobo_atman} proposed a method to manipulate clothing products using a memory gate where the problem is defined as ``fashion search with attribute manipulation". In contrast to \cite{zhaobo_atman}, our proposed method follows a different approach by including the localization of attributes, focusing on attribute representation learning.

\begin{figure*}
\centering
\includegraphics[width=15cm]{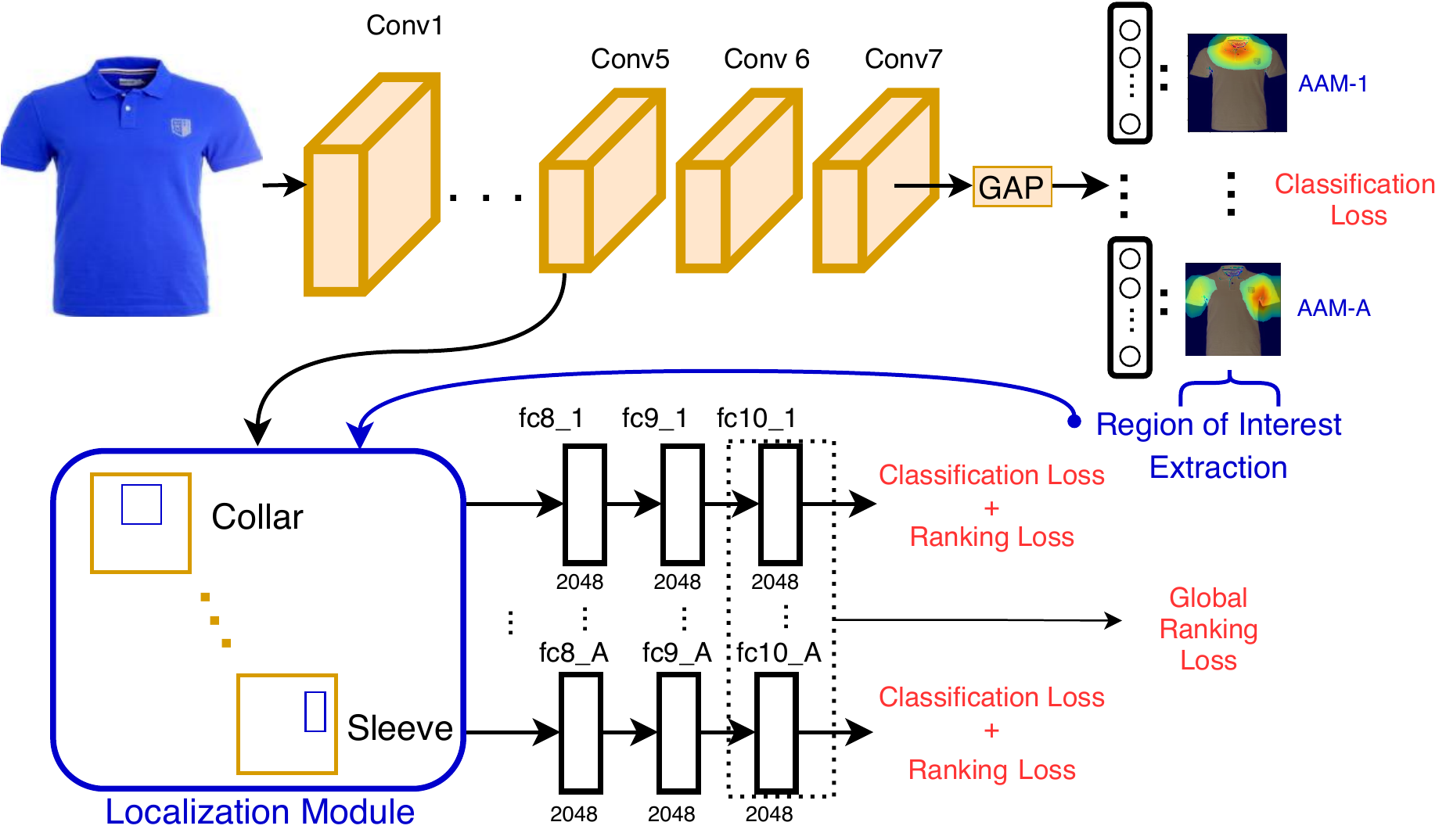}
\caption{Overview of the FashionSearchNet-v2. With the input image fed through the network, the global average pooling (GAP) layer is used to generate attribute activation maps (AAMs) for each attribute which are used to estimate several regions of interests (ROIs). Next, attribute-specific features are pooled from the $conv5$ layer by using the estimated ROIs. The pooled features are linked to a set of fully connected layers where the similarity learning between attributes is conducted and attribute representations are produced ($fc_{10\_1}, fc_{10\_2}, ...$). Finally,  attribute representations are combined into a global representation to be used in the fashion search.}
\label{fig:FashionSearchNet-v2}
\end{figure*} 

\section{FashionSearchNet-v2}
This section presents an overview of the proposed network, FashionSearchNet-v2. First, the network is trained with the classification loss to generate AAMs \cite{zhou2016learning}. Our motivation for using AAMs is to discover the most informative regions for each attribute concurrently, thus ignoring unrelated features. The network is then trained once more with a combination of triplet and classification losses to learn attribute representations with the help of estimated AAMs. Finally, a weighted combination of the attribute representations into a global representation is performed through the global ranking loss. The global representation is computed from the fusion of attribute representations with and without the localization module. The attribute memory block is also updated throughout the training to estimate prototype attribute descriptions to be used in the attribute manipulation step.

\subsection{Architecture Overview}\label{sect:architecture}
The proposed FashionSearchNet-v2 shown in Figure \ref{fig:FashionSearchNet-v2} is based on the AlexNet \cite{Krizhevsky} architecture with the following modifications applied to the baseline network: all fully connected layers are removed and two convolutional layers are added after the $conv5$ layer to compensate the effect of removing the fully connected layers. In the first stage, the network is trained with classification loss to learn accurate AAMs, which represent the most activated regions of attributes. Next, regions of interests (ROIs) are extracted by using the AAMs. In order to learn local attribute representations, feature maps from the $conv5$ layer are passed into a set of fully connected layers with and without ROI pooling. 
In the second stage, the network is trained with the joint combination of classification and ranking losses. These learned attribute representations are combined into a global representation to represent the input image and target attributes. In the final stage, the global ranking loss is applied to estimate the importance of attribute representations depending on the attribute manipulation and learn attribute prototypes.

\subsection{Learning Attribute Representations}\label{sect:LAR}
\textbf{Attribute Activation Maps.} The classification loss is used to discover the most relevant regions of attributes. Initially, the GAP layer is applied to the last convolutional layer which corresponds to $conv7$ as follows:
\begin{equation}
x_{I, k} = \sum_{i,j}conv7_k(I,i,j) \text{ for } k \in 1, 2, ..., K
\end{equation}
where $x_{I, k}$ is the features extracted from the image $I$ for channel $k$ and $conv7_k(I,i,j)$ is the $k$'th feature map of the $conv7$'th layer at spatial location $(i,j)$. The multi-attribute classification network is trained using the following classification loss:
\begin{equation}\label{eq:L_C}
L_{C}  = -{\sum_{I=1}^{N}}{\sum_{a=1}^{A}} \text{log}(p(g_{Ia}|x_{I}w_{a}))
\end{equation}
where $g_{Ia}$ represents the ground truth of the $a'th$ attribute for image $I$. $x_I w_a$\footnote{The dimensions of $w_a$ is [number of feature maps, number of classes associated with $a$]} calculates weighted linear combination of $x_I$ for attribute $a$, $N$ is the number of training examples and $A$ is the number of attributes. The posterior probability estimates the probability of $x_{I}w_{a}$ to be classified as $g_{Ia}$. We next define $M_{a_c}(I,i,j)$ as AAM for class $c$ of an attribute $a$ as follows:

\begin{equation}
M_{a_c}(I,i,j) = \sum_{k}{w_{a_{(k,c)}}conv7_k(I,i,j)}
\end{equation}
where $w_{a_{(k,c)}}$ is the weight variable of attribute $a$ associated with $k'th$ feature map of class $c$ and $c$ is determined from the class, which maximizes the classification confidence. Using $M_{a_c}$, attribute localization can be added to the network. In order to do so, $A$ number of maps are estimated with a simple hard threshold technique. As per the implementation in \cite{zhou2016learning}, the pixel values that are above 20$\%$ of the maximum value in the generated map are segmented. This is followed by estimating a bounding box, that encloses the largest connected region in the AAM. This step is repeated for each attribute. \\
\textbf{Ranking with Triplets of Regions.} FashionSearchNet-v2's ability to identify ROIs enables it to ignore regions with unrelated features which may confuse the attribute similarity learning capability of the network. A structure similar to ROI pooling layer \cite{girshick2015fast} is used to pass features from the $conv5$ layer to a set of fully connected layers.

\begin{figure} 
\centering
\includegraphics[width=8cm]{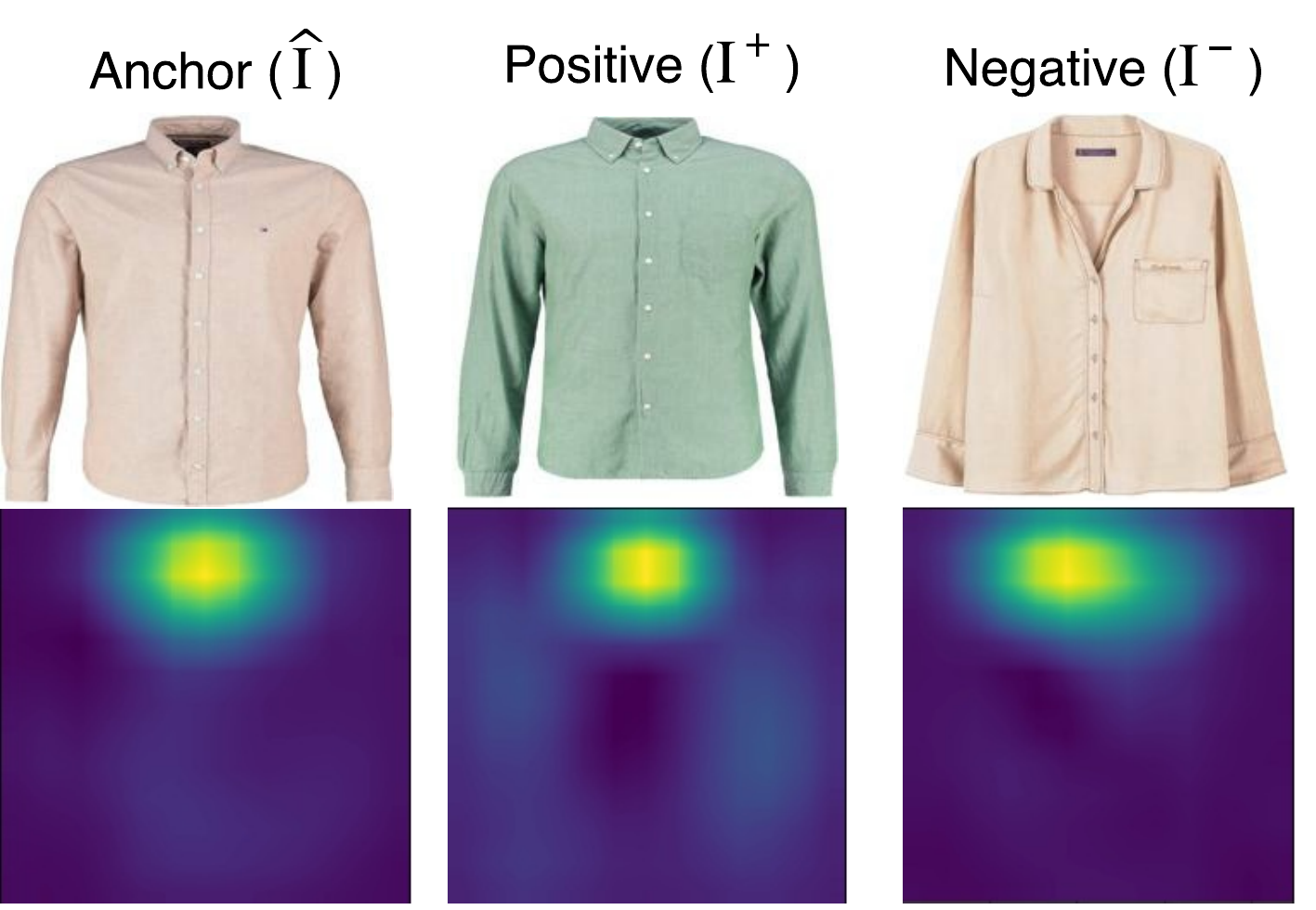}
\caption{Examples for the triplets of regions of the collar attribute: Anchor $(\hat{I})$, Positive $(I^+)$ and Negative $(I^-)$. The generated collar attribute activation maps tend to be near the collar region and irrelevant regions are eliminated; thus, enabling a better attribute similarity learning.}
\label{fig:triplets}
\end{figure}

The example in Figure \ref{fig:triplets} investigates an example for learning collar attribute similarity to show the intuition behind the triplets of regions ranking loss function. At the first glance, anchor image $(\hat{I})$ may look similar to the negative image $(I^-)$, due to color attribute correlation. In fact, the collar attribute of $\hat{I}$ is the same as $I^+$. If the output of the network's $h$'th layer without triplet ranking training is used to compare Euclidean distances, $d(h(\hat{I}),h(I^+)) > d(h(\hat{I}),h(I^-))$ would be the case, meaning $\hat{I}$ would be closer to $I^-$ than $I^+$ in the feature space, which undesired.

The first step of the proposed method involves estimating the corresponding AAMs as shown in Figure \ref{fig:triplets}. Note that the heatmaps of $\hat{I}$  and $I^+$ cover a smaller area compared to $I^-$ thus confirming the localization ability since the collar attribute of $I^-$ covers a wider region. It is evident that as the AAMs localize towards the collar attribute, the unrelated regions such as sleeve and torso are ignored without any intervention. Thus, FashionSearchNet-v2 is able to differentiate the collar attribute while ignoring irrelevant attributes (e.g., color, pattern, etc.).

When the triplet ranking loss function defined in \cite{Huang2015a,shankar2017deep} is used in FashionSearchNet-v2, the observed loss was tremendous unless an extremely small learning rate is used. Inspired by \cite{hoffer2015deep}, the soft-triplet ranking function is utilized which normalizes the distances to the range of (0, 1) with the softmax function and formulated as follows:
\begin{equation}
\scriptstyle
d^+(h(\hat{I}),h(I^+),h(I^-)) = \dfrac{\scriptstyle \text{exp}(||h(\hat{I}) - h(I^+)||_2)} {\scriptstyle \text{exp}(||h(\hat{I}) - h(I^+)||_2) + \text{exp}(||h(\hat{I}) - h(I^-)||_2)} 
\end{equation}
\begin{equation}
\scriptstyle
d^-(h(\hat{I}),h(I^+),h(I^-)) = \dfrac{\scriptstyle \text{exp}(||h(\hat{I}) - h(I^-)||_2)} {\scriptstyle \text{exp}(||h(\hat{I}) - h(I^+)||_2) + \text{exp}(||h(\hat{I}) - h(I^-)||_2)}
\end{equation}
Given $||d^+,d^--1||_2^2 $ $= const.(d^+)^2$ and $h=fc_{10\_a}$ the ranking loss function can be written as:
\begin{equation} \label{eq:L_t}
L_{T} = {\sum_{I=1}^{N}}{\sum_{a=1}^{A}}d^+(fc_{10\_a}(\hat{I}),fc_{10\_a} (I^+),fc_{10\_a} (I^-))
\end{equation}
where A is the number of fully connected layers, which is also equal to the number of attributes. The role of Eq. \ref{eq:L_t} is to learn a representation for each attribute using the final set of fully connected layers: $fc_{10\_a}$. We minimize $||fc_{10\_a}(\hat{I}),fc_{10\_a}(I^+)||_2$ and maximize $||fc_{10\_a}(\hat{I}),fc_{10\_a}(I^-)||$. The rule for picking triplets is as follows:, $\hat{I}$ and $I^+$ must share the same label while $I^-$ is chosen randomly from a different label. For instance, given if an anchor includes ``blue" color label, the positive image can be any image with ``blue" color label.

Both ranking and classification losses are used in the optimization processes leading to the attribute representations. It is necessary to use the classification loss as it was observed from experiments that using only the ranking loss significantly diminishes the discriminative ability of the network. This classification loss denoted as $L_{TC}$ is formulated as in Eq. \ref{eq:L_C}, except $x_{I}w_{a}$ is replaced with the output of $fc_{10\_a}$ layers. 
\\
\textbf{Feature Fusion.} Compared to the previous version, i.e., FashionSearchNet \cite{ak2018learning}, we also include global image features from $conv5$ layer when computing feature representations by concatenating features from localized \& global feature maps before the first fully connected layer.

\subsection{Attribute Manipulation \& Learning Global Representation} \label{sect:learn_global}
In the previous subsection, we showed how FashionSearchNet-v2 is taught to localize and learn attribute representations. Combining all these learned features would achieve good results when conducting image searches. However, such combinations may allocate too much memory and thus slow down the search process. Incorporating additional training also helps the network to learn how to merge feature representations for attribute manipulation.

By associating each attribute with a different fully connected layer ($fc_{10\_1},  ..., fc_{10\_A}$), the fashion search with attribute manipulation becomes straightforward. After the training, features with the same attribute value are extracted from the training images and stored in a local representation block $M$ $\in$ $\mathbb{R}^{C \times D}$ where $C$ is the total number of attribute values and $D$ is the feature dimension.

\begin{figure}
\centering
\includegraphics[width=8.5cm]{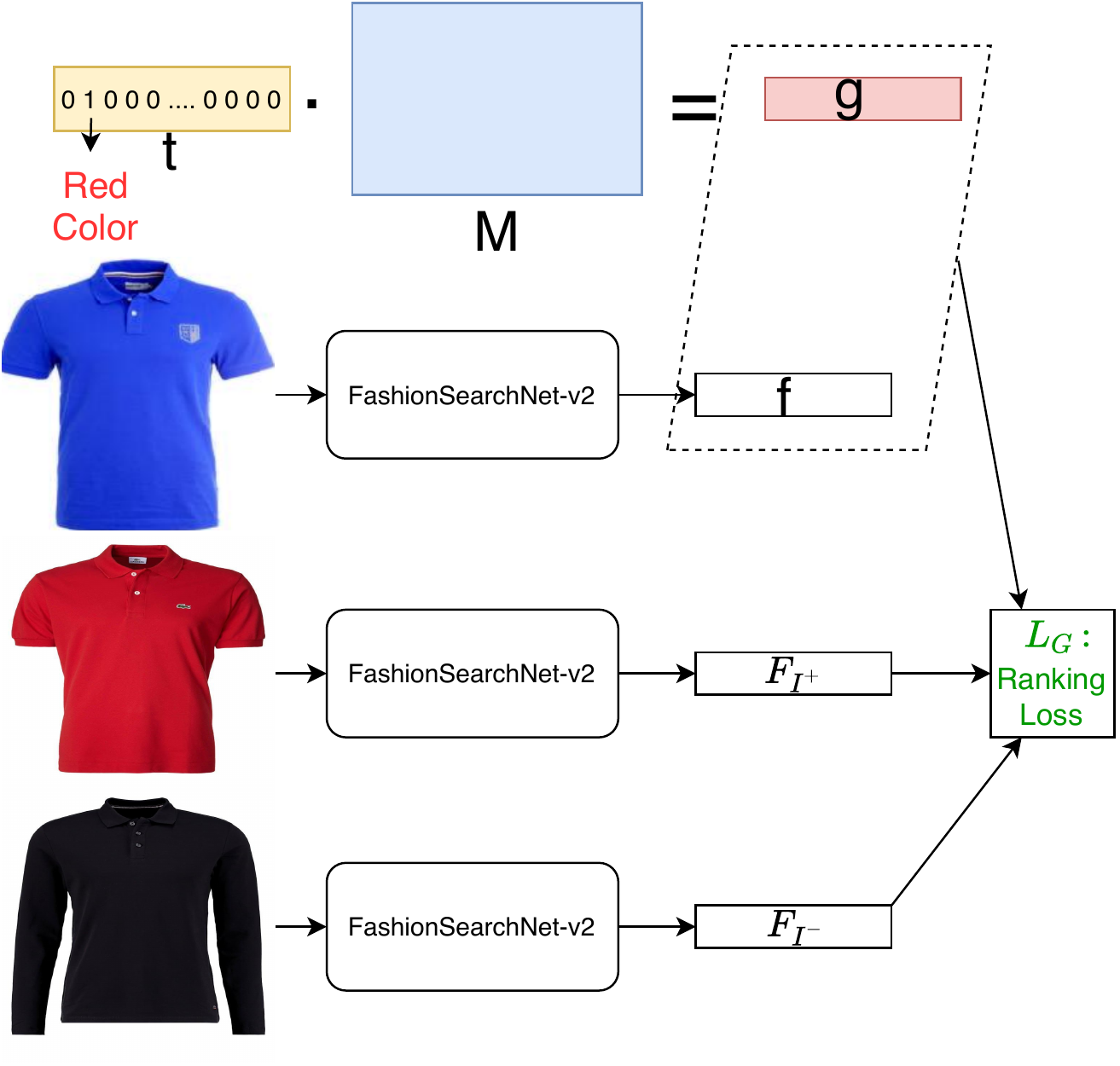}
\caption{Attribute retrieval module example. Given a query image and attribute manipulation (red color), the network is trained with a triplet loss where the positive image is the one matches all attributes and the negative image is chosen randomly.}
\label{fig:attribute_retrieval_module}
\end{figure}

Given an attribute manipulation $t \in \mathbb{R}^{1 \times C}$, the corresponding attribute representation can be retrieved via $g=tM$ as visualized in Figure \ref{fig:attribute_retrieval_module}. After retrieving the new representation $g$, it is combined with feature representation of the query image $f$ and the undesired representation of the query image is overthrown. This formulation also enables the FashionSearchNet-v2 to update features in $M$, which would improve the retrieval performance.

In order to reduce the dimension of the concatenated feature representation of $(f, g)$, a weight parameter $w_{r}$ is applied to reduce the concatenated feature-length to the dimension $r$. Moreover, we use an additional ranking function and letting a weight variable learn ``which features are more important" to boost the performance vastly. This is because some attribute representations such as ``color" could be more important than other attribute representations say, ``fabric". $A$ number of weight parameters denoted by $\lambda _{a}$ are learned. The training is conducted with the following global ranking loss function, $L_G$ for a given attribute manipulation ${t}$, which replaces attribute $a$:
\begin{equation}\label{eq:F_IT}
F_{I,t} =  [fc_{10\_1}(I)\lambda_{1},..., g\lambda_a, ...,fc_{10\_A}(I)\lambda_A]w_{{a^*}}
\end{equation}
\begin{equation}\label{eq:L_G}
L_{G}  = {\sum_{I=1}^{N}}d^+(F_{I, t} ,F_{I^+} ,F_{I^-})
\end{equation}
where $F_{I^+}$ and $F_{I^-}$ are features of positive and negative triplet samples, respectively. For the training of global representations, the unwanted attribute representation of the query image is replaced with $g$, which corresponds to $F_{I, t}$ and weight variables applied as shown in Eq. \ref{eq:F_IT}. The training is conducted with the loss function given in Eq. \ref{eq:L_G}. The same procedure is applied to all possible attribute manipulation options. The rule for picking triplets for the global ranking loss is that $F_{I,t}$ and $F_{I^+}$ must be identical in terms of attributes after the attribute manipulation $t$ while $F_{I^-}$ is chosen randomly.

Another advantage of global representation learning is not only weights are updated but also attribute representations in $M$ as all operations are differentiable. Following such updates on the memory block allows the network to optimize its parameters to find the most representative features for an attribute manipulation.


\subsection{Optimization}
FashionSearchNet-v2 utilizes different loss functions in its optimization steps. The joint loss can be written as follows:
\begin{equation}\label{eq:joint}
L_{joint} = \lambda_{C}L_{C} + \lambda_{T}L_{T} + \lambda_{TC}L_{TC} + \lambda_{G}L_{G}
\end{equation}

The network is first trained only using $L_{C}$ as the other processes depend on how reliable the AAMs are. In the second stage, we assign the following weights: $\lambda_{C}=1, \lambda_{T}=1.5, \lambda_{TC}=1, \lambda_{G}=0$ to further train the network. After the training is finished, memory block $M$ is constructed and another training is performed for global representation learning using only $L_{G}$ where local attribute representations are kept fixed. Note that, it is possible to perform another joint training but due to memory efficiency we used fixed features for the global ranking loss.

\section{Experiments}
\subsection{Implementation Details}
We use pre-trained ImageNet weights for AlexNet \cite{Krizhevsky} up until $conv4 'th$ layer and reinitialize other layers. For $conv5$, $conv6$, $conv7$ layers, sizes of feature maps are 384, 512, 512, respectively. As we use regions of triplets ranking constraint, the selection of $(\hat{I}, I^+, I^-)$ becomes easy. For each mini-batch, images with the same attribute are chosen to be $\hat{I}$ and $I^+$. $I^-$ is picked such that it has a different attribute. Gradients are calculated for each loss function and they are backpropagated through the network. 

The network is trained with the stochastic gradient descent algorithm using the learning rate of $0.01$. No pre-pro-cessing steps other than removing the means from each channel of images were conducted. For the ROI pooling, Tensorflow's \cite{abadi2016tensorflow} ``tf.image.crop\_and\_resize'' function is used to feed the $conv5$ features with the estimated bounding boxes into a set of fully connected layers. Dropout was used on all fully connected layers with $p = 0.5$. For all datasets, networks are trained for 12 epochs for the first stage, 12 epochs for the second stage (joint loss), and 2 epochs for global ranking loss.

\begin{table*}[ht]
\caption{Top-20 retrieval accuracy for each available attribute in their respective datasets.}
\centering
\begin{subtable}[h]{\textwidth}
\centering
\begin{tabular}{c c c c c c c c c c c c c | c}
\hline
Approach & Category & Color  & Collar & Fabric & Fasten. & Fit & Gender & Neckline  & Pocket & Pattern & Sleeve  & Sport &  \textbf{Avg.} \\  \hline
AMNet & 0.135 & 0.278 & 0.326 & 0.351 & 0.232 & 0.393 & 0.092 & 0.204 & 0.278 & 0.304 & 0.096 & 0.227 & 0.243 \\ \hline
w/o Rank  & 0.085 & 0.271 & 0.223 & 0.317 & 0.199 & 0.343 & 0.118 & 0.139 & 0.222 & 0.422 & 0.101 & 0.199 & 0.220
 \\ 
Rank  & 0.141 & 0.306 & 0.379 & 0.357 & 0.258 & 0.396 & 0.200 & 0.177 & 0.292 & 0.490 & 0.240 & 0.215 & 0.288  \\ 
Rank-L  & 0.168 & 0.410 & 0.431 & 0.390 & 0.265 & 0.398 & 0.249 & 0.222 & 0.335 & 0.461 & 0.334 & 0.247 & 0.326
 \\
Rank-LG  & 0.350 & \textbf{0.611} & 0.605 & 0.403 & 0.344 & 0.489 & 0.426 & 0.476 & 0.519 & 0.510 & 0.563 & 0.374 & 0.473
  \\ 
Full & 0.336 & 0.569 & \textbf{0.613} & 0.395 & \textbf{0.364} & \textbf{0.510} & \textbf{0.446} & \textbf{0.491} & \textbf{0.521} & \textbf{0.549} & \textbf{0.590} & \textbf{0.410} & \textbf{0.483}
 \\
Full w/ FF & \textbf{0.369} & 0.549 & 0.599 & \textbf{0.405} & \textbf{0.364} & 0.470 & 0.429 & 0.447 & 0.504 & 0.500 & 0.565 & 0.275 & 0.456
  \\ \hline
\end{tabular}
\caption{Shopping100k dataset.}
\hfill
\end{subtable}

\begin{subtable}[h]{\textwidth}
\centering
\begin{tabular}{c c c c c c c c c c | c}
\hline
Approach & Button & Category  & Collar & Color & Length & Pattern & Shape &  Sleeve Len. & Sleeve Shp. & \textbf{Avg.} \\ \hline
AMNet  & 0.253 & 0.183 & 0.191 & 0.202 & 0.185 & 0.168 & 0.205 & 0.100 & 0.173 & 0.184    
\\ \hline
w/o Rank  & 0.192 & 0.097 & 0.156 & 0.110 & 0.115 & 0.123 & 0.124 & 0.127 & 0.099 & 0.127   
\\ 
Rank & 0.262 & 0.162 & 0.337 & 0.227 & 0.152 & 0.202 & 0.198 & 0.159 & 0.148 & 0.205   
\\ 
Rank-L  & 0.353 & 0.330 & 0.412 & 0.354 & 0.258 & 0.276 & 0.308 & 0.245 & 0.249 & 0.309
   \\ 
Rank-LG  & \textbf{0.401} & 0.353 & \textbf{0.414} & 0.391 & 0.363 & \textbf{0.298} & \textbf{0.377} & 0.359 & 0.247 & 0.356
    \\ 
Full  & 0.398 & \textbf{0.388} & 0.387 & 0.398 & \textbf{0.369} & 0.280 & 0.355 & 0.327 & 0.252 & 0.350
    \\
Full w/ FF  & 0.383 & 0.350 & 0.408 & \textbf{0.406} & 0.365 & 0.287 & 0.374 & \textbf{0.432} & \textbf{0.252} & \textbf{0.362}
   \\ \hline 
\end{tabular}
\caption{DARN dataset}
\hfill
\end{subtable}

\begin{subtable}[h]{\textwidth}
\centering
\begin{tabular}{c c c c c c c c c | c}
\hline
Approach & Category & Color  & Gender & Material & Neckline & Pattern & Sleeve &  Style & \textbf{Avg.} \\ \hline
AMNet  & 0.039 & 0.067 & 0.111 & 0.050 & 0.046 & 0.071 & 0.062 & 0.038 & 0.061

      \\ \hline 
w/o Rank   & 0.024 & 0.037 & 0.095 & 0.038 & 0.049 & 0.028 & 0.017 & 0.025 & 0.039

    \\ 
Rank    & 0.043 & 0.065 & 0.124 & 0.055 & 0.065 & 0.057 & 0.047 & 0.056 & 0.064

    \\ 
Rank-L    & 0.048 & 0.098 & 0.111 & 0.066 & 0.073 & 0.092 & 0.074 & 0.060 & 0.078

    \\ 
Rank-LG  & 0.076 & 0.138 & \textbf{0.160} & 0.080 & 0.091 & 0.114 & 0.082 & 0.059 & 0.100

   \\ 
Full  & \textbf{0.093} & \textbf{0.144} & 0.137 & \textbf{0.102} & \textbf{0.104} & \textbf{0.131} & 0.091 & \textbf{0.070} & \textbf{0.109} \\
Full w/ FF   & 0.088 & 0.134 & 0.154 & 0.094 & 0.096 & 0.129 & \textbf{0.092} & \textbf{0.070} & 0.107
 \\  \hline
\hline
\end{tabular}

\caption{iMaterialist dataset}
\hfill
\end{subtable}

\begin{subtable}[h]{\textwidth}
\centering
\begin{tabular}{c c c c c c c c c | c}
\hline
Approach & Hair Color & Beard  & Hair Type & Smiling & Eyeglass & Gender & Hat &  Age & \textbf{Avg.} \\ \hline
AMNet  & 0.604 & 0.439 & 0.556 & 0.496 & 0.275 & 0.293 & 0.155 & 0.370 & 0.399
      \\ \hline 
w/o Rank  & 0.468 & 0.290 & 0.530 & 0.563 & 0.163 & 0.315 & 0.127 & 0.396 & 0.357
    \\ 
Rank   & 0.816 & 0.500 & 0.710 & 0.805 & 0.747 & 0.782 & 0.642 & 0.515 & 0.690
    \\ 
Rank-L  & 0.818 & 0.556 & 0.752 & 0.806 & 0.779 & 0.812 & 0.748 & 0.537 & 0.726
    \\ 
Rank-LG  & 0.836 & 0.598 & \textbf{0.788} & 0.809 & 0.785 & 0.818 & \textbf{0.769} & 0.759 & 0.770
   \\ 
Full  & \textbf{0.841} & 0.604 & 0.787 & \textbf{0.815} & 0.785 & \textbf{0.829} & 0.766 & \textbf{0.772} & \textbf{0.775}
    \\ 
Full w/ FF  & 0.827 & \textbf{0.613} & 0.783 & 0.805 & \textbf{0.789} & 0.813 & \textbf{0.771} & 0.766 & 0.771
    \\  \hline
\end{tabular}

\caption{CelebA dataset}
\end{subtable}
\label{tab:top_20_table}
\end{table*}

\begin{figure*}
\centering
\begin{subfigure}[b]{0.495\textwidth}
\includegraphics[width=\textwidth]{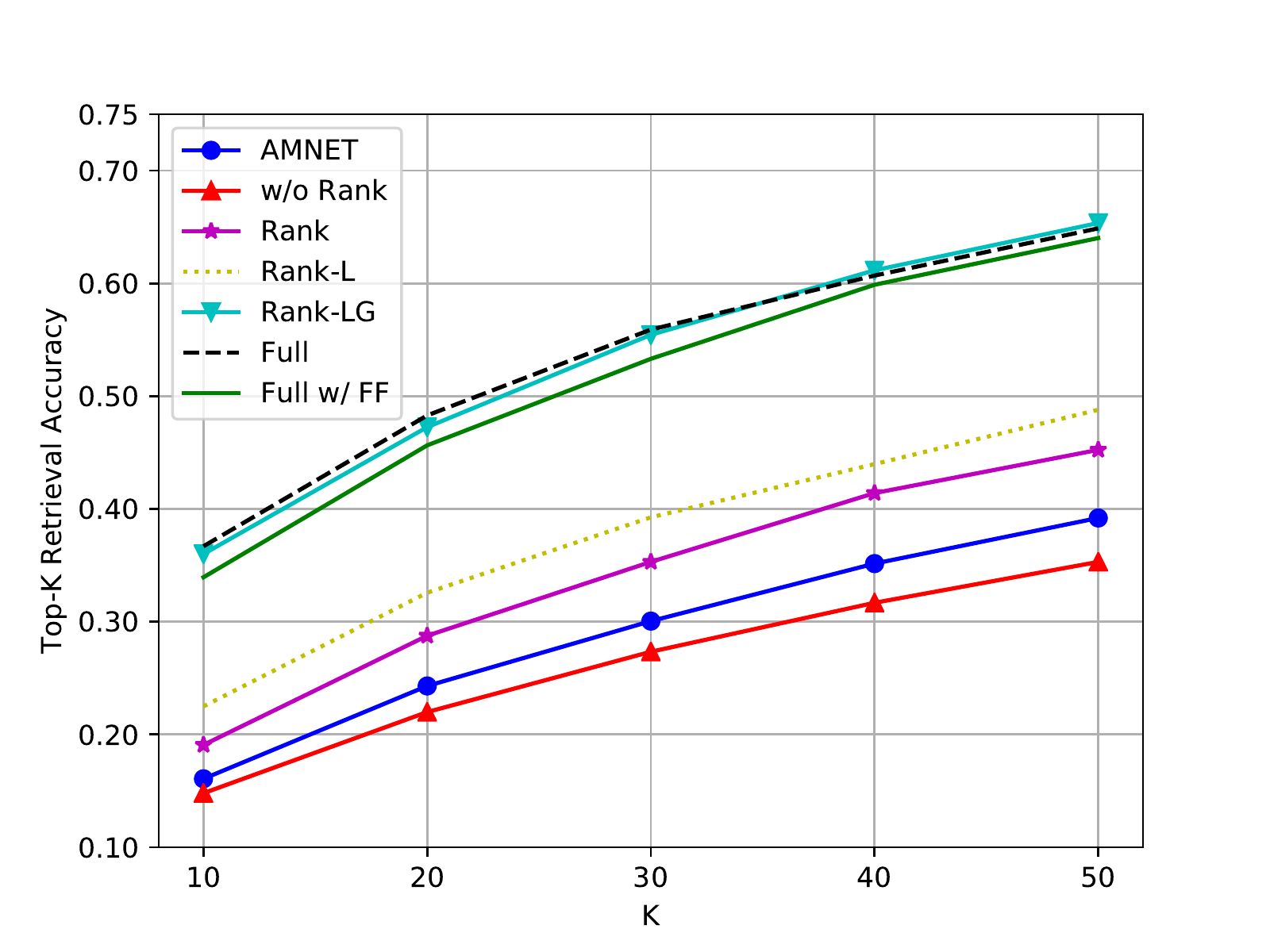}
\caption{Shopping100k}
\end{subfigure}
\begin{subfigure}[b]{0.495\textwidth}
\includegraphics[width=\textwidth]{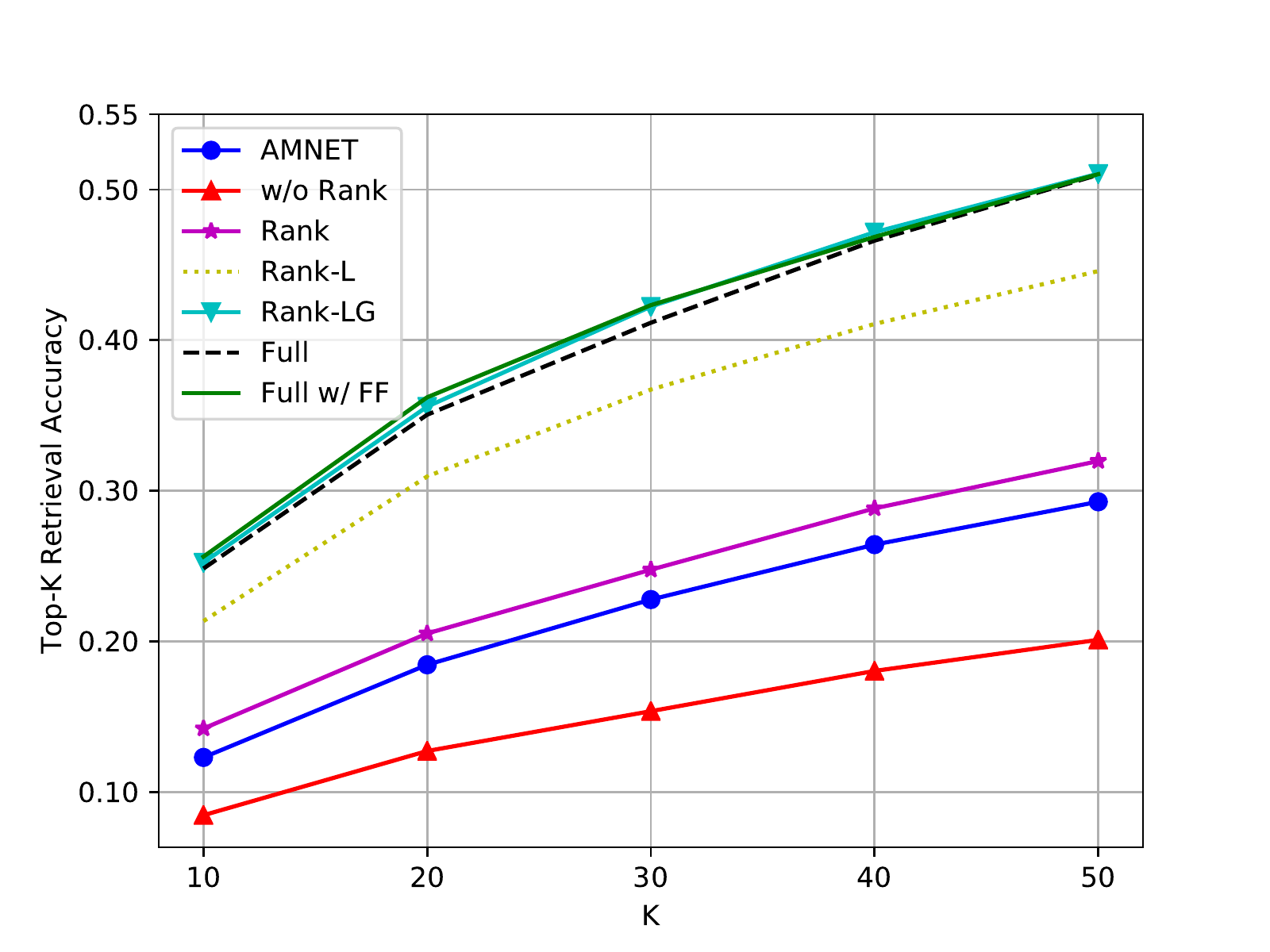}
\caption{DARN}
\end{subfigure}
\begin{subfigure}[b]{0.495\textwidth}
\includegraphics[width=\textwidth]{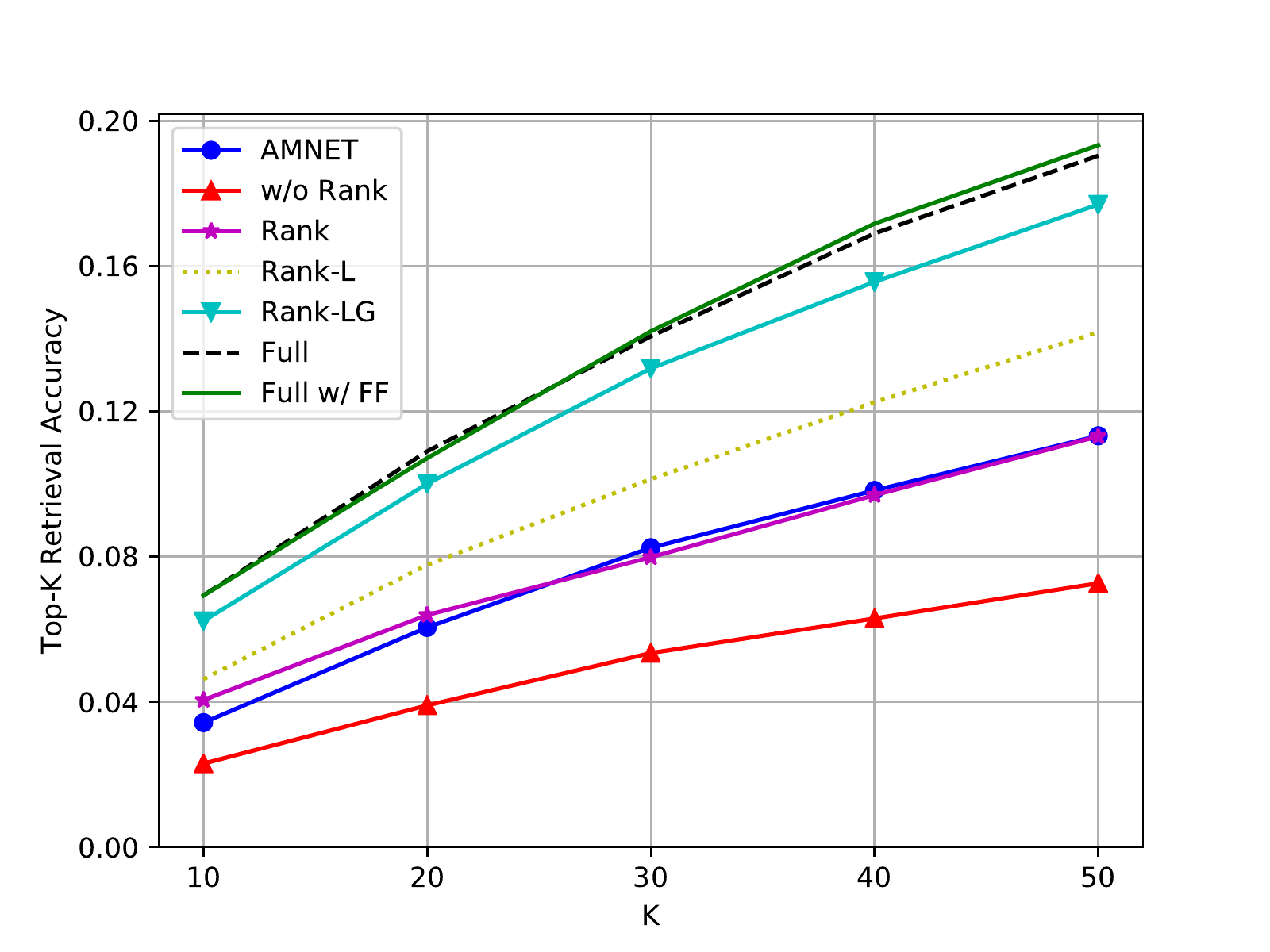}
\caption{iMaterialist}
\end{subfigure}
\begin{subfigure}[b]{0.495\textwidth}
\includegraphics[width=\textwidth]{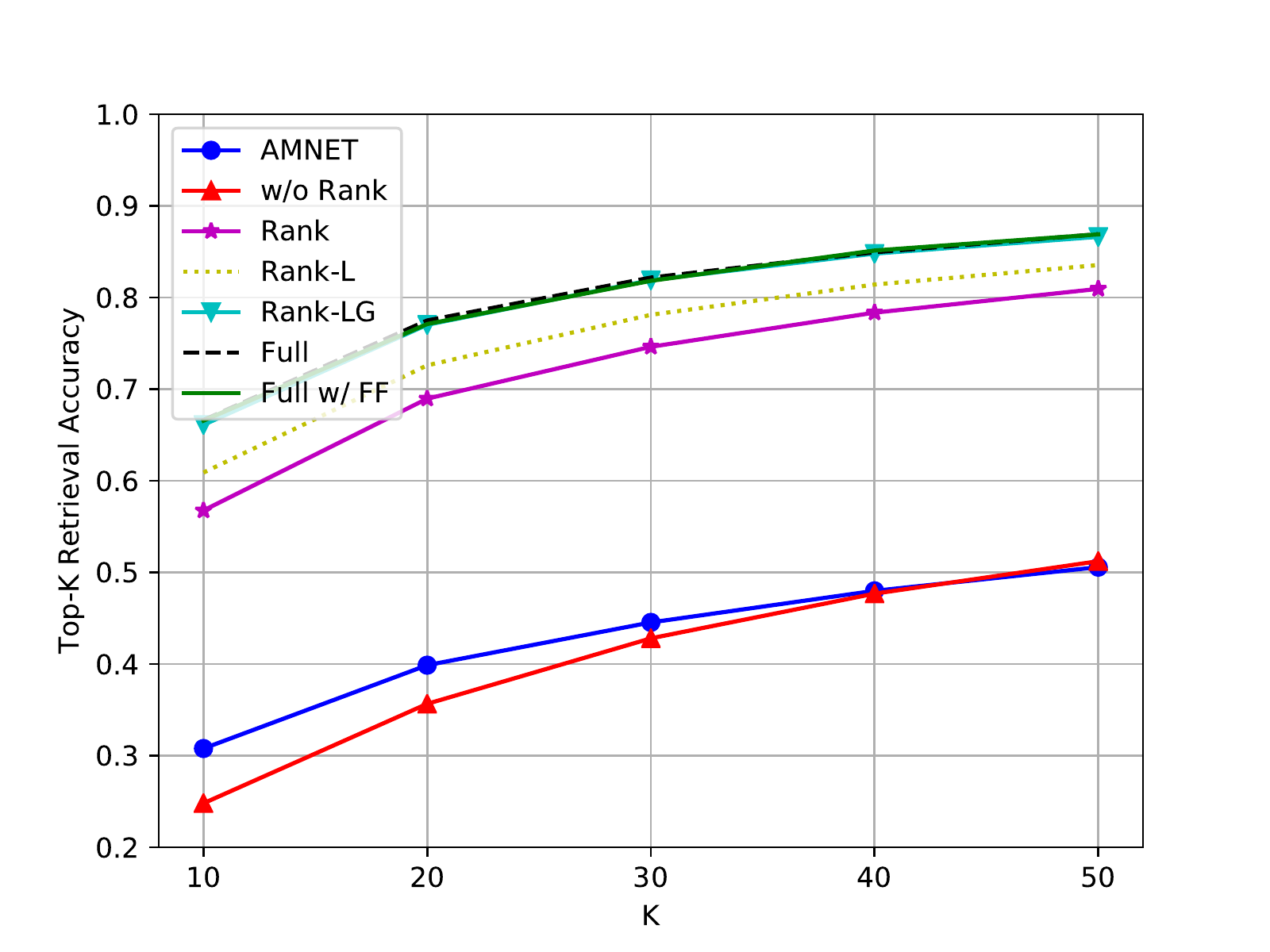}
\caption{CelebA}
\end{subfigure}
\caption{Top-K retrieval accuracy with varying K value for attirbute manipulation experiments in (a) Shopping100k, (b) DARN, (c) iMaterialist and (d) CelebA datasets.}
\label{fig:all_plots}
\end{figure*}

\subsection{Datasets}
Of the many datasets available that include fashion images and attributes \cite{ak2018efficient, hadi2015buy,Huang2015a,kiapour2014hipster,Liu,jia2020fashionpedia}, we decided to use Shopping100k \cite{ak2018efficient}, DARN \cite{Huang2015a} and iMaterialist \cite{guo2019imaterialist} datasets in our experiments. These fashion datasets cover a large variety of attributes and are rich in the number of images. Additionally, we use the CelebA dataset \cite{liu2015faceattributes} to show that the proposed method can be generalized to non-fashion images too. In all our experiments, we use 2,000 images to serve as query and 18,000 images as retrieval gallery, the rest are used for training the networks. Some details of these datasets listed as follows:
\begin{itemize}
    \item \textbf{Shopping100k} dataset contains 101,021 images with 12 attributes and in total 151 unique values are available. Different than the others, this dataset does not include any person in the image where only the clothing product with a simple background is available.
    \item \textbf{DARN} dataset includes 272,711 fashion images with 9 clothing attributes and the total number of attribute values is 179.
    \item \textbf{iMaterialist} dataset has around 1 million fashion images. After removing images with noisy labels, we select a subset of 250,000 images to be used in our experiments. We also group some similar category labels together reducing the number of unique category labels from 228 to 24. In total there are 8 attributes with 147 attribute values.
    \item \textbf{CelebFaces Attributes (CelebA)} dataset is a large-scale face attributes dataset with 202,599 number of face images where each image has 40 binary attributes. We group these attributes as hair-color, beard, hair-type, smiling, eyeglasses, gender, hat, and age for our experiments leading to 8 attributes with 21 attribute values.
\end{itemize}
\subsection{Competing Methods}
We investigate several state-of-the-art approaches and found that FashionNet \cite{Liu} and StyleNet \cite{simo2016fashion} which solve different fashion retrieval problems and are not suitable for attribute manipulation. We compare the performance of FashionSearchNet-v2 with its earlier version \cite{ak2018learning} and AMNet \cite{zhaobo_atman}.

We additionally use different variations of the proposed method in ablation experiments to investigate the effect of each novel component such as attribute localization, global representation learning, updatable memory block and feature fusion.
\subsection{Evaluation Metric} 
For the qualitative experiments, we define the retrieval accuracy as follows. Given a query image and an attribute manipulation, the search algorithm finds the ``best K" image matches i.e., ``Top-K" matches. If there is a match (i.e., having the same attributes as the query image) after attribute manipulation, it corresponds to a hit ($1$) otherwise it is a miss ($0$).
\begin{figure*}[t]
    \centering
    \includegraphics[width=\textwidth]{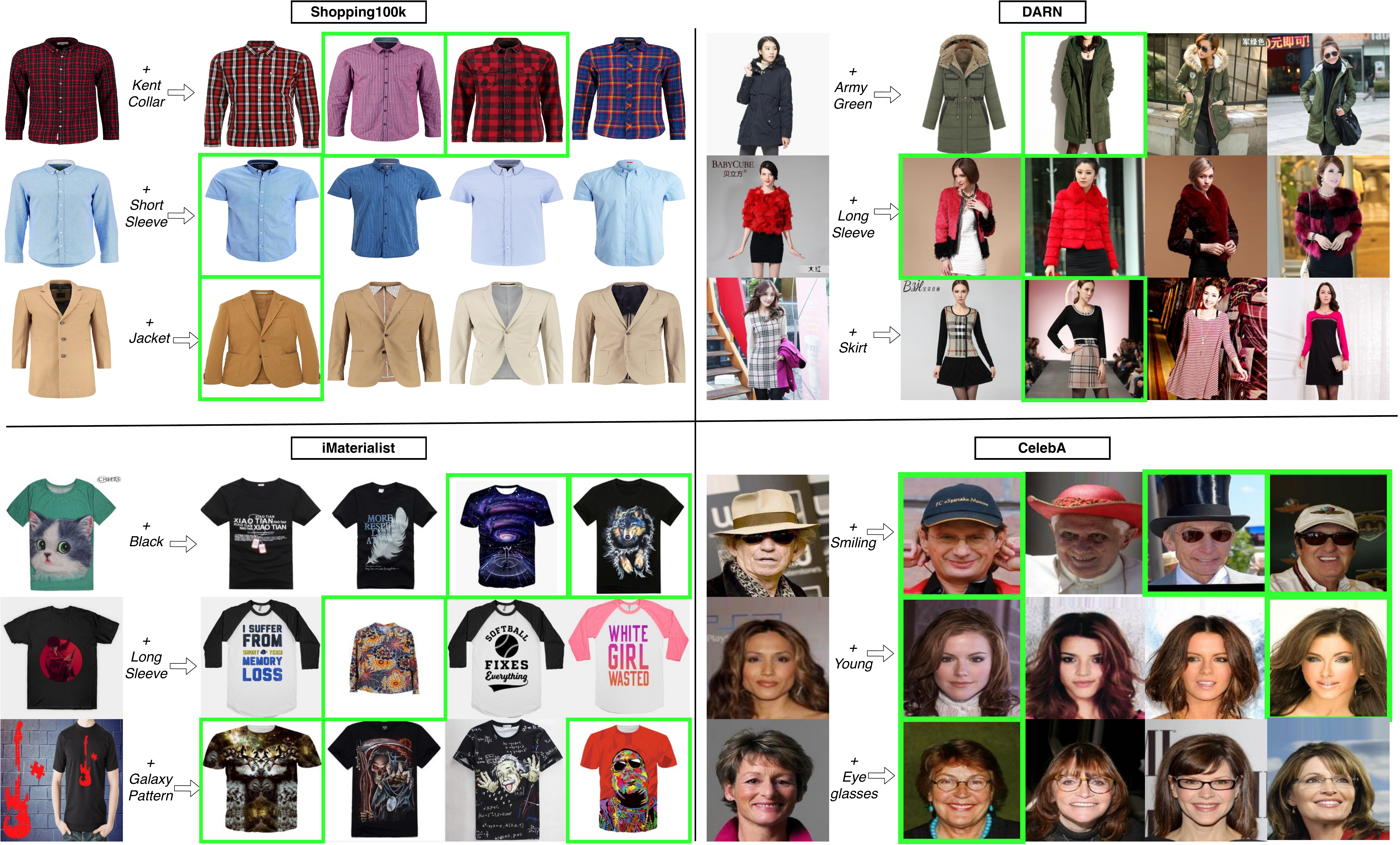}
    \caption{The Top-4 retrieval results of fashion search with attribute manipulation for a given a query image for all datasets. The green bounding boxes denote images retrieved by FashionSearchNet-v2 match all desired attributes when the search is conducted after the attribute manipulations.}
\label{fig:all_examples}
\end{figure*}
\\
\textbf{Search Strategy. }As discussed in \cite{Datta2008}, the search involves the use of very precise terms resulting in short search sessions leading to an end-result. We also adopt simple fashion queries by only changing a single attribute from the query image.
%
\subsection{Attribute Manipulation Experiments}
Our experiments involve replacing certain feature representations of the query image with the desired and comparing Euclidean distances of global representations with the retrieval gallery. For these experiments, every possible attribute manipulation that is available in the retrieval gallery is applied to the query images.

Top-K retrieval accuracy results are presented in Figure \ref{fig:all_plots} and Table \ref{tab:top_20_table} for (a) Shopping100k, (b) DARN, (c) iMaterialist and (d) CelebA datasets, respectively. FashionSearchNet-v2 achieves the best performance throughtout all experiments. Compared to the only other available baseline \textit{AMNet}, it can be seen in Table \ref{tab:top_20_table} and Figure \ref{fig:all_plots}, there is a large disperency compared to the variations of our proposed methods: \textit{Full} and \textit{Full w/ FF}. The main reason for this performance gap is the lack of attribute specific learning and localization of attributes in \textit{AMNet}. Next, we perform an ablation study to investigate different aspects of the proposed method.

\textbf{Ranking Loss. }To obtain insights about the proposed architecture, we start our experiments from simpler baselines and build on top of each preceding model. \textit{w/o Rank} model is based on AlexNet with additional final fully connected layers to match the number of attributes and does not incorporate any ranking loss, trained with the classification loss. For attribute manipulation, we use a technique to directly replace the matching representation from the learned attribute memory $M$. Compared to the \textit{w/o Rank} model, the inclusion of the ranking loss \textit{Rank} brings out a significant performance boost in all experiments as shown in Table \ref{tab:top_20_table} and Figure \ref{fig:all_plots}.

\textbf{Localization. }Next, we extend the \textit{Rank} model and incorporate the AAMs to the framework: \textit{Rank-L}, which provides the localization mechanism. We observe that \textit{Rank-L} significantly outperforms \textit{Rank} throughout all datasets in Table \ref{tab:top_20_table}, indicating the importance of localization. Looking at Figure \ref{fig:all_plots}, the contribution of the localization module is much more significant in the DARN dataset than Shopping100k. This is mostly due to more complicated images as seen in Figure \ref{fig:all_examples} where it is harder to estimate good attribute representations. 

\textbf{Global Representation Learning. }In order to learn the importance of attribute representations, we introduced a technique in Sect. \ref{sect:learn_global}. By incorporating this into \textit{Rank-L} , we construct \textit{Rank-LG} where we observe a great improvement in Top-K retrieval accuracy for all datasets as shown in Table \ref{tab:top_20_table} and Figure \ref{fig:all_plots}. This technique is significant to determine which features are more important during attribute manipulation. \textit{Rank-LG} corresponds the earlier version of FashionSearchNet-v2 \cite{ak2018learning}.

\textbf{Updatable Memory Block. }In \textit{Rank-LG}, the memory $M$ is fixed the same during the global representation learning. Since the retrieval of attribute representation from the memory block is a differential operation, we can also update features in $M$. We denote this model as \textit{Full}, which achieves improvements over \textit{Rank-LG} but the margin increase is mostly not as significant as the other proposed techniques. Improvements are observed in Table \ref{tab:top_20_table}, except the DARN dataset.

\textbf{Feature Fusion. }In our final experiments, we also include Feature Fusion (FF) to \textit{Full}, which results to \textit{Full w/ FF}. The inclusion of feature fusion is important when the attribute localization method makes mistakes and in that case, the addition of the whole feature map can be helpful to recover the issue caused by the incorrect localization. Feature fusion achieves some improvements but the overall performance is close to the \textit{Full} model. This study also shows that the localized attribute representations from the localization mechanism are generally correct and feature fusion does not result in a significant improvement.

With regard to the fashion datasets, it can be seen that the overall performance in Shopping100k is greater than the other datasets. This is mostly due to simple, clean images of fashion items that enable the models to learn faster. In the iMaterialist dataset, the results are much lower compared to DARN where both datasets have people wearing fashion items. This is mostly due to noisy street pictures in the iMaterialist dataset. 

Lastly, in the CelebA dataset, the overall performance is much higher as face pictures are taken in frontal and the number of attributes is smaller than those of the fashion datasets. Additionally, the facial attributes are not as complicated compared to the fashion datasets.
\subsection{Qualitative Results}
Figure \ref{fig:all_examples} shows several examples of the fashion search experiments for all datasets used in experiments. Images with the green bounding box mean that all attributes are matched with the query image and attribute manipulation. In most cases, the proposed algorithm can retrieve several relevant images with various attribute manipulations.

For the Shopping100k dataset, images do not have any wearer in them, which makes it easier than the other two fashion datasets due to more clean shot photos. We provide several examples in Figure \ref{fig:all_examples} wherein each row, the proposed method can retrieve images with the desired attribute manipulation. In unsuccessful cases, although the desired attribute is included in the retrieved images, the other attributes of the query image get affected by the attribute manipulation, which leads to the retrieval of wrong images.

For the DARN dataset, there are more various images compared to the Shopping100k dataset in terms of pose and lighting variations. The proposed method can handle both global (+army green) and local (+long sleeve) attribute manipulations. In the final row, a more challenging query formulation (dress to skirt) is where the proposed method can also handle.

The iMaterialist dataset is more difficult as Top-K retrieval accuracy in overall is much lower than the other datasets according to Table \ref{tab:top_20_table} and Figure \ref{fig:all_plots}. This is mostly due to the number of attributes and images are taken from different perspectives. We show several successful examples in Figure \ref{fig:all_examples}.

Lastly, in the CelebA dataset, the proposed method is mostly successful and can retrieve many relevant images given the query image and attribute manipulation. In the final row, the method can retrieve only one image that matches the conditions however other images also seem visually correct. This may be due to the labeling of the dataset where two similar people may be labeled slightly different.

\subsection{Run Time Performance}
Our FashionSearchNet-v2 is trained on Intel i7-5820K CPU and 64 GB RAM memory with GeForce GTX TITAN X GPU. FashionSearchNet-v2 can extract features from 10,000 images in around 60 seconds which is very close to the attribute-based method. Compared to the AlexNet implementation \cite{Krizhevsky}, the proposed FashionSearchNet-v2 has several additional layers to learn attribute representations. However, by using smaller fully connected layers, the run-time performance of FashionSearchNet-v2 is still efficient compared to the attribute-based method. Moreover, using the ROI pooling layer, all images just need to be fed into the network once which saves a lot of computation time. The extraction of ROIs for all attributes is efficient as it takes about only 0.002 seconds for each image.

\section{Conclusion}
This paper presents a new approach for conducting fashion searches using query images and attribute manipulations. The proposed FashionSearchNet-v2 is able to estimate efficient feature representations for fashion search and its good localization ability enables it to identify the most relevant regions for attributes of interest. In addition to being able to combine attribute representations into a single global representation for attribute manipulations, FashionSearchNet-v2 incorporates different techniques such as memory update and feature fusion. The proposed FashionSearchNet-v2 is shown to outperform the baseline fashion search methods including AMNet \cite{zhaobo_atman}. An interesting problem for future work would be to extend FashionSearchNet-v2 for more flexible attribute queries and comparative operations.

\bibliography{MyCollection}
\bibliographystyle{IEEEtran}

\end{document}